\documentclass[twocolumn]{article}
\usepackage[a4paper, total={6.5in, 9in}]{geometry}

\usepackage{booktabs}
\usepackage{epsfig}
\usepackage{subcaption}
\usepackage{calc}
\usepackage{amssymb}
\usepackage{amstext}
\usepackage{amsmath}
\usepackage{amsthm}
\usepackage{multicol}
\usepackage{pslatex}
\usepackage{apalike}
\usepackage{algorithm2e}
\usepackage[bottom]{footmisc}

\theoremstyle{definition}
\newtheorem{definition}{Definition}

\newcommand{\R}{\mathbb{R}}
\renewcommand{\P}{\mathbb{P}}

\newcommand{\X}{\mathcal{X}}

\newcommand{\T}{\mathcal{T}}
\newcommand{\D}{\mathcal{D}}

\usepackage{tikz}
\usetikzlibrary{positioning,shapes}
\usepackage{hyperref}
\usepackage{cleveref}
\renewcommand*{\thefootnote}{\fnsymbol{footnote}}
\usepackage{url}
\title{Investigating the Suitability of Concept Drift Detection for Detecting Leakages in Water Distribution Networks}
\author{Valerie Vaquet\footnote{Autors contributed equally},\, Fabian Hinder,$^*$ and Barbara Hammer}
\date{Machine Learning Group, Bielefeld University, Germany}
\begin{document}

\maketitle




\renewcommand*{\thefootnote}{\arabic{footnote}}
\abstract{Leakages are a major risk in water distribution networks as they cause water loss and increase contamination risks. Leakage detection is a difficult task due to the complex dynamics of water distribution networks. In particular, small leakages are hard to detect. From a machine-learning perspective, leakages can be modeled as concept drift. Thus, a wide variety of drift detection schemes seems to be a suitable choice for detecting leakages. In this work, we explore the potential of model-loss-based and distribution-based drift detection methods to tackle leakage detection. We additionally discuss the issue of temporal dependencies in the data and propose a way to cope with it when applying distribution-based detection. We evaluate different methods systematically for leakages of different sizes and detection times. Additionally, we propose a first drift-detection-based technique for localizing leakages.}


\section{{Introduction}}
\label{sec:introduction}
Clean and safe drinking water is a scarce resource in many areas. Almost 80\% of the world's population is classified as having high levels of threat in water security~\cite{vorosmarty2010global}. This will aggravate in the future as due to climate change the already limited water resources will become more restricted~\cite{rodell2018emerging}.
Currently, across Europe, considerable amounts of drinking water are lost due to leakages in the system\footnote{\url{https://www.eureau.org/resources/publications/1460-eureau-data-report-2017-1/file}}. The issue has been identified by the European Union, which put the topic on the political agenda recently\footnote{\url{https://eur-lex.europa.eu/legal-content/EN/TXT/HTML/?uri=CELEX:32020L2184&from=ES##d1e40-1-1}}.

In order to ensure a reliable drinking water supply, there is a need for reliable, safe, and efficient water distribution networks (WDNs). In addition to avoiding water losses, a crucial requirement is to ensure the quality of the drinking water, e.g. to avoid the spreading and growth of bacteria and other contaminants. Leakages pose a major risk to water quality as they make it possible for unwanted substances to enter the water system. Thus, monitoring the system for leakages is an efficient tool to avoid water loss and contamination~\cite{eliades_fault_2010,lambert_accounting_1994}.

Due to complex network dynamics and changing demand patterns detecting both small and large leakages is a challenging task. This is aggravated by the fact that the available data is very limited. Usually, the precise network topology remains unknown or the documentation contains errors. As smart meter technologies are not widely distributed there is no real-time demand information~\cite{smart-meters}. In realistic settings, this leaves a set of scarce pressure and possibly flow measurements. 


Commonly, existing leakage detection methodologies rely on replicating the system of interest by means of hydraulic models and monitoring the discrepancies between observations and modeled values. While these approaches can provide reasonable detection when considering larger leakages, the approaches struggle when facing smaller leakages~\cite{vrachimis_battle_2022}. Besides, there are major problems hindering the usage of these applications in real-world applications. First of all, rebuilding a hydraulic system requires detailed information about the network layout, e.g. exact pipe diameters, pipe roughness, and elevation levels. Frequently, this kind of information is not available to water supply companies, as the pipe infrastructure has been in place for a long time. Second, hydraulic simulations require real-time demands as an input for the computations. This is not a realistic requirement as the availability of smart metering systems is still limited due to the associated cost and privacy concerns~\cite{smart-meters}. Finally, the need to adapt the simulation arises whenever the system is changing. 
Next to the hydraulic approaches, there are also a few machine learning-based approaches that implement a similar strategy.

In this work, we focus on the problem of leakage detection from the perspective of handling data streams containing temporal dependencies. More precisely, we formalize leakages as concept drift and the problem of leakage detection as drift detection. Our aim is to investigate the suitability of drift detection for reliable leakage detection, whereby we focus on leakage of all practically relevant sizes. Our approach is independent of the specific WDN and requires only real-time pressure measurements. Thus, it is more flexible and efficient in comparison to the family of hydraulic simulation-based approaches.

This paper is structured as follows. First, we introduce WDNs and summarize the main specifics of this domain (\cref{sec:wdn}). Afterward, we briefly summarize the body of related work on leakage detection (\cref{sec:rel-work}). In \cref{sec:drift}, we define concept drift and cover both model-loss-based and distribution-based drift detection. Before evaluating the suitability of these methodologies for leakage detection in \cref{sec:experiments}, we discuss the issue of temporal dependencies in the data collected from WDNs and propose ways to account for those (\cref{sec:dependencies}). Finally, we conclude our paper in \cref{sec:conclusion}.

\section{Water Distribution Networks\label{sec:wdn}}

WDNs can be modeled as graphs consisting of nodes representing junctions and undirected edges representing pipes as the flow direction of the water is not pre-defined and can change over time due to different inputs and demands in the network. As usually real-world water distribution systems are extended over time, it is possible that multiple pipes lie parallel to each other \cite{mala-jetmarova_lost_2018}. As the systems are observed over time, hydraulic quantities like pressure and flow describe the network graph at each time step. They can be described by hydraulic formulas given that the network is known in great detail. Next to the exact pipe layout, different parameters like elevations, pipe diameters, and pipe roughness are required. Assuming sensors are installed in $n$ nodes across the graph, for each time step $t$, measurements $x_t=[x_1^t \dots x_n^t]$ are collected where $x_i^t$ is the value at node $i$. 

Real-world measurements of the complete hydraulic state of WDNs are not available as it is not possible to measure the entire system. Even a precise topology alongside measurements in many positions is usually not available since making this information public would be a weakness of the system which is part of the critical infrastructure. Even if this data were available, for the task of leakage detection, no ground truth of when exactly a leakage occurred at which position would be available. Thus, when developing and evaluating methodologies monitoring WDNs one usually relies on simulated data. Given the key parameters of a system (layout, elevations, pipe information) alongside demand patterns realistic network states containing anomalies like leakages can be simulated using simulation tools like EPANET~\cite{rossman_epanet_2000}.

As WDNs are part of the critical infrastructure and are required to work robustly and safely to ensure the health and well-being of the population, additional requirements are put on monitoring tools, especially those using AI technologies. Besides requirements concerning robustness, safety, and fairness as formulated in the European AI-Act~\cite{EU_proposal_AI},
some technical attributes of WDNs pose additional challenges to ML approaches. When working with WDNs, only limited knowledge about the pipe system is available. Usually, the exact properties of the pipes, e.g. their diameters and roughness, and the different elevation levels are unknown. In some cases, even the precise pipe layout is not documented reliably~\cite{eggimann_potential_2017}. Note that these are available for a few benchmark networks, and thus benchmark scenarios can be generated. However, when designing monitoring systems relying on this kind of information strongly limits the applicability in practical applications. As we will discuss in the next section, the lack of this kind of knowledge limits the application of many state-of-the-art solutions.

Besides limited information about the system setup, the system is also relatively opaque with respect to the real-time dynamics. Due to installation costs and challenges regarding the power supply, the availability of pressure and flow sensors in WDNs is very limited yielding readings at a fraction of the nodes in the system. Data availability is even more limited for real-time demand measurements, as households are very rarely equipped with smart meters for drinking water due to costs and data privacy.

Another property of WDNs is the presence of cycling patterns in demands, flows, and pressures. When working on ML approaches, one needs to account for the presence of temporal dependencies. As the water demands of both industrial and private customers are not constant overall the demands fluctuate over the course of a day. Additionally, the demand patterns vary between work days and weekends, and changes caused by different seasons and long-term developments, e.g. climate change or the COVID pandemic, need to be expected. These fluctuations increase the difficulty of leakage detection as especially smaller leakages might be lost in the signals.

\section{Related Work\label{sec:rel-work}}
The body of related work on leakage detection can be divided into methods relying on a hydraulic model and very few ML-based approaches. Hydraulic model-based methods generally aim to replicate the real-world system with a hydraulic model~\cite{hu_survey_2018}. Usually, the simulation results of the hydraulic model are then compared to the observations. An anomaly is reported if the residual of these methods is too large, which is determined either by a threshold~\cite{romero-ben_leak_2022}, a CUSUM approach~\cite{steffelbauer_pressure-leak_2022}, or visual inspection~\cite{marzola_leakage_2022}. All these methods share the downside that they require real-time demands and more information on the network topology than is usually available~\cite{vrachimis_battle_2022}. Besides, they lack generalizability across WDNs as the hydraulic model is specifically designed for one network and even needs adaptation if something changes within this particular network. While these hydraulic-based approaches yield good results considering large leakages they usually miss smaller ones~\cite{vrachimis_battle_2022}.

There are few ML-based approaches for leakage detection~\cite{daniel_sequential_2022,laucelli_detecting_2016,romano_automated_2014}. However, many are only evaluated on very small networks and/or lack realistic demands as input for the simulation data they are tested on. Most of these approaches replace the hydraulic model with some ML model following the same general idea of residual-based anomaly detection, for example by using a threshold~\cite{daniel_sequential_2022,laucelli_detecting_2016}.

\section{Detecting Concept Drift\label{sec:drift}}

Deploying machine learning-based systems in real-world scenarios, one needs to account for all kinds of changes and ensure that the models reliably work even if the observed environment changes. Thus, considerable research focuses on ML in the presence of changes in the data-generating process, which are called concept drift or drift for shorthand. In order to obtain a formal definition of drift, we first need to define a so-called drift process\cite{hinder2023things}:
\begin{definition}
    Let $\T = [0,1]$ and $\X = \R^d$.
    A \emph{drift process} $(P_T,\D_t)$ from the \emph{time domain} $\T$ to the \emph{data space} $\X$ is a probability measure $P_T$ on $\T$ together with a Markov kernel $\D_t$ from $\T$ to $\X$, i.e., for all $t \in \T$ $\D_t$ is a probability measure on $\X$ and for all measurable $A \subset \X$ the map $t \mapsto \D_t(A)$ is measurable. We will just write $\D_t$ instead of $(P_T,\D_t)$ if this does not lead to confusion.
\end{definition}
Based on this a definition of drift can be obtained:
\begin{definition}
    Let $(P_T,\D_t)$ be a drift process. We say that $\D_t$ has \emph{drift} iff
    \begin{align*}
    \P_{T,S \sim P_T}[\D_T \neq \D_S] = P_T^2(\{(t,s) \in \T^2 \mid \D_t \neq \D_s \}) > 0.
\end{align*}
\end{definition}

Next to a branch of research focusing on the task of keeping an ML model accurate in the presence of drift (\emph{online or stream learning}), another branch focuses on \emph{monitoring and analyzing data streams} with respect to drift. In this work, we will focus on the latter as we are interested in the monitoring task of leakage detection.

Drift detection schemes can be generally categorized into \emph{model-loss-based} and \emph{distribution-based} approaches. While the latter directly investigates the observed data, model-loss-based approaches first train a model and then analyze its loss as a proxy for change in the data distribution. Here, the rationale is that a drift event changes the data in a way that the model cannot approximate well anymore, causing a decline in the model loss. This approach is valid if used in online learning where the goal is to keep a model which performs well at its prediction task. Yet, as argued by \cite{hinder_change_2023,hinder_hardness_2023} the relation between model-loss and drift is rather loose -- in case the model does not provide sufficient complexity to approximate the data distribution well (i) the drift might stay undetected as it is smoothed out by the model or in converse (ii) the model might change because of irrelevant drifts, e.g. a change in the ratio of classes, causing a change in the model-loss even though no real drift occurred. For these reasons, when working on monitoring tasks, one should apply distribution-based approaches. 

However, frequently model-loss-based approaches are also used in monitoring tasks. 
For instance, the residual-based strategy described in \cref{sec:rel-work} which is used across most state-of-the-art approaches published on leakage detection can be categorized as model-loss-based approach. Therefore, we will investigate the suitability of both types of drift detection methods in this work.

\subsection{Model-loss-based Drift detection\label{sec:lossdd}}
Performing model loss-based drift detection the idea is to analyze the model loss as a proxy for drift in the data distribution. In case of WDNs, there are two reasonable inference tasks a model can perform as a proxy for the prediction: Either one performs a \emph{forecasting} task where the goal is to predict the measurement of next time step $x_{t+1}$ based on the sensor measurements collected up to time $t$, or one performs an \emph{interpolation} task where the goal is to predict one sensor by the measurements of all other sensors, i.e. for each node position $i$, a model $f_i:\R^{n-1}\rightarrow\R,\,f_i(x^t_{\setminus i})=\hat{x}^t_{i}$ is trained. $x^t_{\setminus i}$ only includes the measurements of the other sensors $\{s_j\}_{j\in\{1,\dots n\},\,j\neq i}$ at time $t$. The latter strategy has been employed as a virtual sensor imputation strategy in case of sensor faults. Even very simple ML models could successfully perform the interpolation task~\cite{vaquet_taking_2022}.

\subsection{Distribution-based Drift detection\label{sec:distdd}}
Most distribution-based approaches follow the strategy of \emph{comparing two samples}. 
This can be done by \emph{statistical testing}: Considering two windows some test is applied to determine whether they are generated by the same distribution. One particular popular test is the \emph{Kolmogorov-Smirnow (KS) test}~\cite{an1933sulla}. However, since this test is only working on one-dimensional data, a feature-wise testing scheme is required. In contrast, the \emph{kernel two-sample test} which relies on the maximum mean discrepancy (MMD) can test distributions of a higher degree by computing a kernel matrix as a descriptor~\cite{mmd}.

Another option to include vectorial data is to use a \emph{virtual classifier} as a first step. This strategy follows the fact that machine learning classifiers perform better than random guessing if they can exploit some properties of the data. The idea is to consider the two windows as different classes in a classification task. If the obtained model performs better than it would be by chance, the distributions of the windows differ, i.e. a drift occurred. For this methodology, any classifier can be chosen. However, the model class determines the detection capabilities. We will consider the D3 detection scheme~\cite{d3} in our experiments.

A methodology not focusing on two samples but instead relying on \emph{meta-statistics} is the shape drift detector~\cite{hinder2021shape}. It computes the discrepancy of two consecutive time windows which was introduced as the drift magnitude that is sliding through the stream. Plotting those discrepancies over time, in case of drift, this graph will take a particular shape that can be used to detect and pinpoint it. We will not use the ShapeDD directly in our experiments as its main purpose is to increase the performance of the MMD, but we will make use of it for analysis purposes. 

Finally, a \emph{block-based} detection scheme can be used. In contrast to all aforementioned methods, it does not rely on comparing two data samples obtained at different time points to test for drift but rather searches directly for a dependency of data and time which was identified to be an equivalent description of drift by \cite{hinder2020dynamic}. This task can be performed by a standard independence test; in this work, we will make use of the HSIC-test~\cite{hsic} which is another kernel-based method.

As discussed in \cref{sec:wdn}, different kinds of daily, weekly, and seasonal patterns have to be expected. These patterns introduce certain temporal dependencies to the data. As already discussed, these patterns might increase the difficulty of detecting leakages. Considering this from a theoretical viewpoint, this problem can be summarized by the need to account for the temporal dependencies when performing drift detection. 
Thus before, describing the drift detection schemes used in our experiments, in the next section we will analyze the temporal patterns in the data.

\section{Temporal dependencies in the data\label{sec:dependencies}}
We already raised the issue of temporal dependencies in data collected from WDNs. In this section, we will analyze the dataset which we will use in our experiments later on. For this purpose, we will first briefly introduce the dataset and provide our analysis.

\subsection{L-Town Benchmark Data\label{sec:data}}

\begin{figure}
    \centering
    \includegraphics[width=0.5\textwidth]{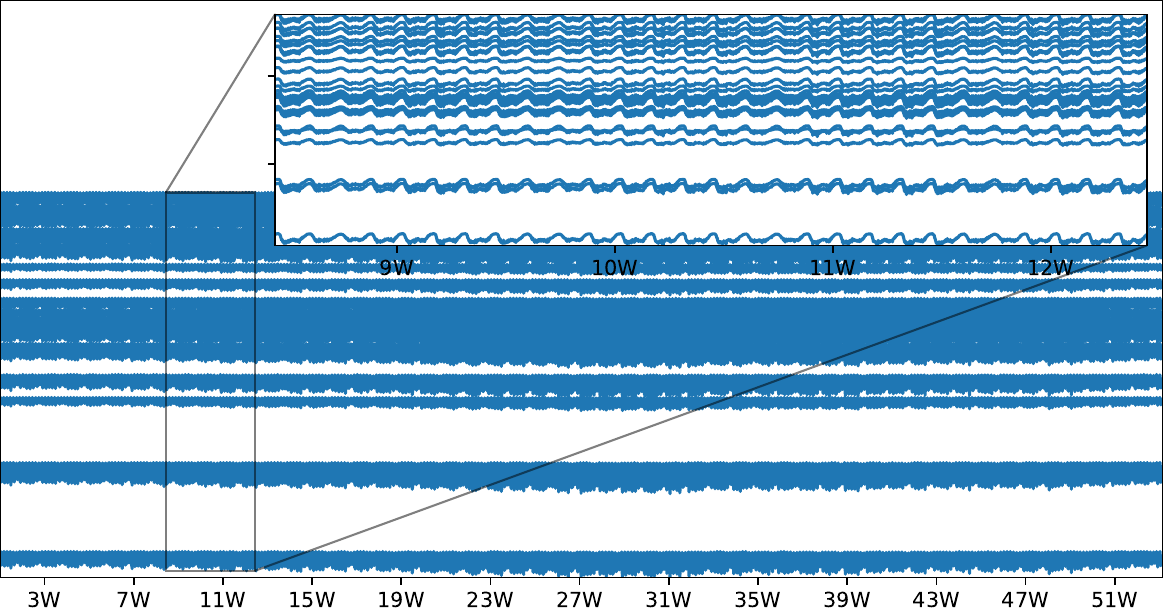}
    \caption{Visualization of sensor data for one year (no leak).}
    \label{fig:raw-data}
\end{figure}

In this work, we will consider the L-Town network since it is relatively complex in comparison to other benchmarks and one year of realistic real-time demands are available for this system allowing us to simulate realistic data for our experimental evaluation. The L-Town network resembles parts of the old town of Limassol, Cyprus. In our experiments, we consider area A consisting of 661 nodes and 764 edges with 29 optimally placed pressure sensors~\cite{vrachimis_battle_2022}. We run simulations with four different leakage sizes ranging from 7mm to 19mm at all pipes using the ATMN package which builds on EPANET. Each scenario contains data of 364 days with a measuring frequency of 15 minutes. We always consider one leakage per scenario which starts at some point of the scenario and stays present until the scenario ends. 

\subsection{Analysis\label{sec:analysis}}
Analyzing the data, as expected we observe daily, weekly, and seasonal patterns. As visualized in \cref{fig:raw-data}, the pressure follows a clear weekly pattern as can be seen in the zoom-in subplot.
In order to control those dependencies we perform two analysis strategies: 1) subtracting the ``standard week'', 2) subtracting the values of the previous week.

\begin{figure}
    \centering
    \includegraphics[width=0.5\textwidth]{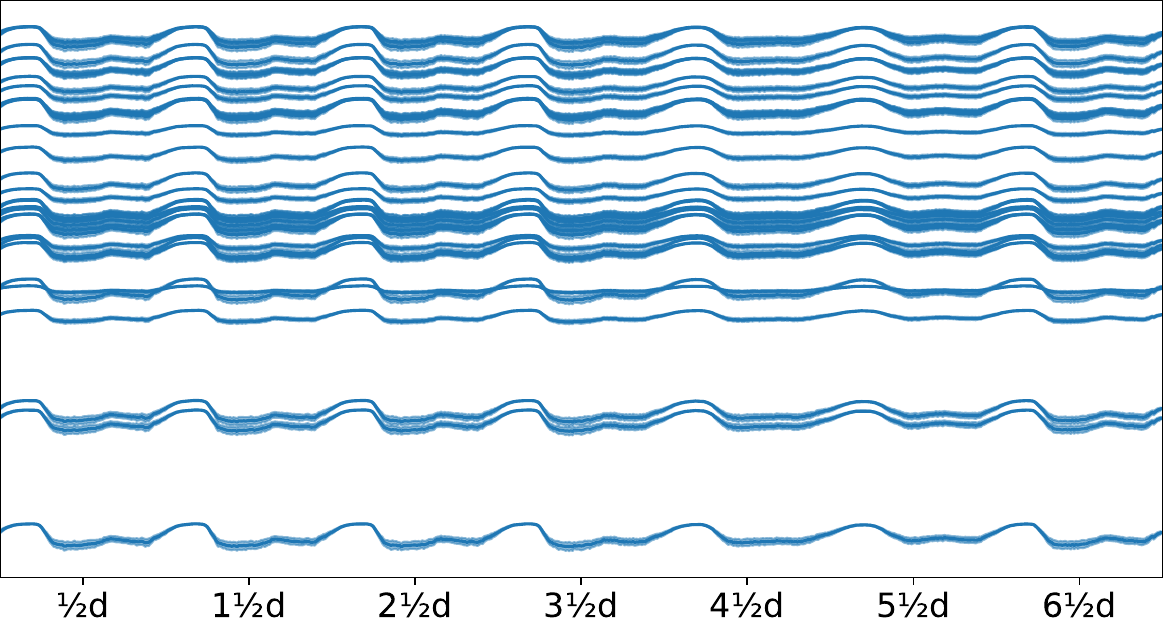}
    \caption{Visualization of standard week (line), shaded area shows standard deviation. }
    \label{fig:std-week-plot}
\end{figure}
The standard week is the average over all weeks. We have illustrated it in \cref{fig:std-week-plot}. As can be seen, the standard deviation is comparably small implying that the weekly patterns are relatively consistent throughout the year. Furthermore, we observe the same strong daily patterns we have observed before as well as a difference in the shape of two days, probably corresponding to weekends. 

\begin{figure}
    \centering
    \includegraphics[width=0.5\textwidth]{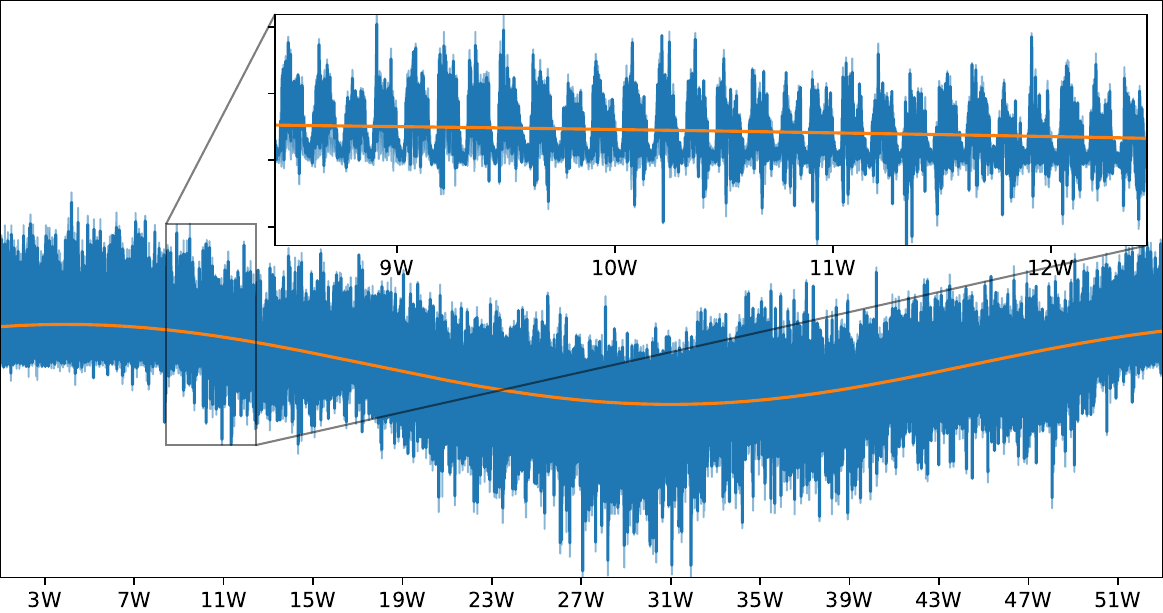}
    \caption{Visualization of sensor residuals after subtracting standard (no leak). Orange line marks mean trend across all sensors.}
    \label{fig:std-week-error}
\end{figure}
By subtracting the standard week from the original signals we obtain the signals shown in \cref{fig:std-week-error}. The plot shows one examplary sensor reading (blue line) as well as the minimal and maximal sensor reading at each given point in time as shaded area. As can be seen, the feature runs across the entire range implying very strong fluctuations. Furthermore, as can be seen in the zoomed-in plot there is a change in fluctuation that follows a daily pattern. 
In addition, we also added a trend line (orange) which follows a cosine shape. This is a plausible finding as we expect a cyclic pattern across several years that correlates with the seasons. However, this pattern may render change detection schemes useless as it induces changes that are not caused by leaks. 

\begin{figure}
    \centering
    \includegraphics[width=0.5\textwidth]{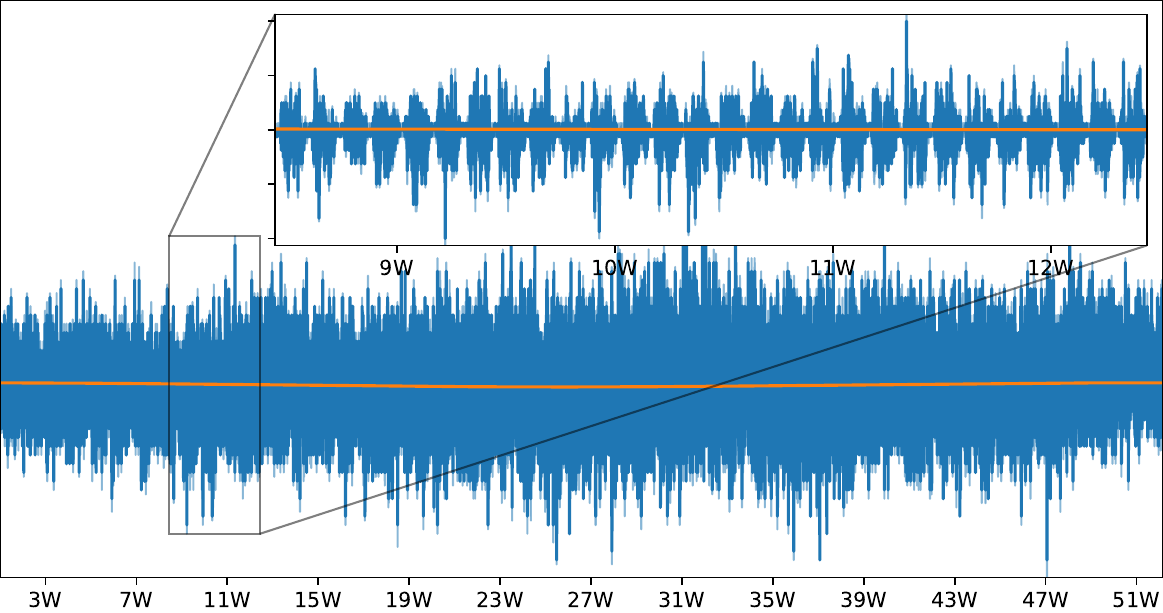}
    \caption{Visualization of sensor residuals after subtracting value of last week (no leak). Orange line marks mean trend across all sensors.}
    \label{fig:week-diff-error}
\end{figure}
As an alternative, we considered subtracting the value of the last week rather than a standard week. The results are illustrated in \cref{fig:week-diff-error}. Due to the small variance in the computation of the standard week, this is already a good proxy for the standard week. However, it is better suited to cope with long-term changes as can be seen from the trend line. Furthermore, we again observe strong oscillations whose intensities follow a daily pattern. We will find that this strategy is actually quite efficient in \cref{sec:model-loss}.

\begin{figure*}[h]
    \centering
    \includegraphics[width=\textwidth]{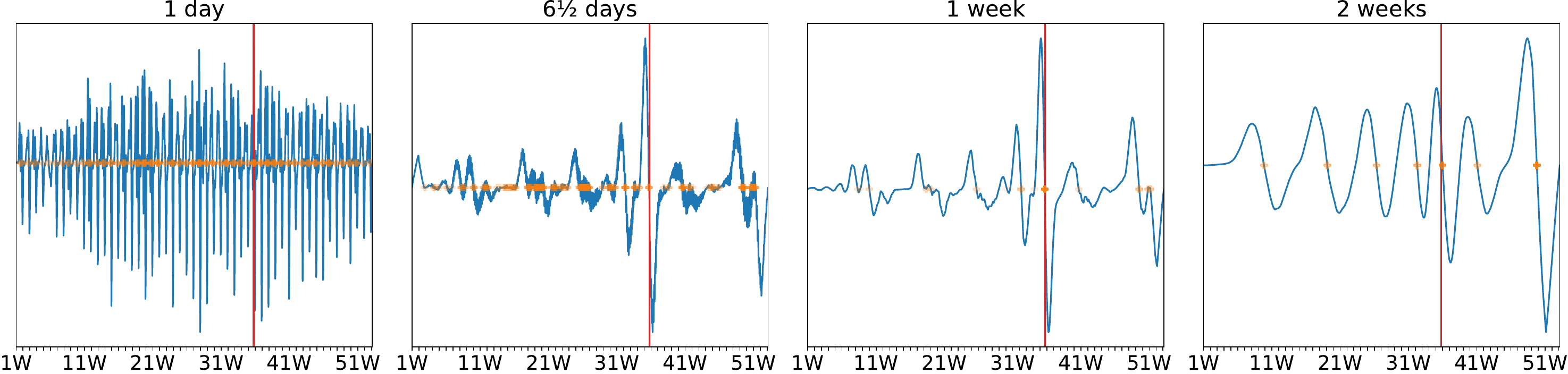}
    \caption{Plot of shape curve for different windows sizes. Red line marks time point of drift, orange crosses mark candidate time points (shape heuristic) with transparency indicating value of MMD at that point.}
    \label{fig:shape}
\end{figure*}

From both analyses, we expect that we can easily cope with the periodic patterns if we only compare the data on a by-week basis. This is due to the fact that there is a noticeable difference between the values of weekends and weekdays so that day-wise is too short to resolve this dependency. Furthermore, longer periods of time will be strongly affected by the seasonal trends. We will further discuss those ideas in the next section.

\subsection{Coping with temporal dependencies in the data}

We observed substantial temporal patterns in the data which we need to account for when utilizing drift detection schemes for the task of detecting leakages. For model-loss-based drift detection approaches we assume that the models can generalize well. Thus, in this setting, no additional actions need to be taken. In contrast, when using distribution-based schemes, we need to carefully incorporate our knowledge of the different temporal cycles in the data to successfully detect leakages. 

In preliminary experiments, we used a pre-processing technique, which subtracted a standard week to eliminate cycles in the data. However, this strategy assumes that we can model this standard week successfully which requires some leakage-free historical data. Since we aim to develop a methodology that requires as little information as possible to generalize to new networks, we additionally experimented with choosing the window sizes such that the detection schemes do not suffer from seasonalities. Here, as discussed before our idea is to eliminate the cyclic patterns by choosing exactly one week per window. Thereby daily and weekly patterns are eliminated while the windows are still small enough to not be affected by long-term dependencies. Since this strategy resulted in better results while requiring no additional information, we will use this option in our experimental evaluation instead of performing a preprocessing step subtracting the standard or the previously observed week.

We analyzed the efficiency of this approach by making use of the Shape Drift Detector~\cite{hinder2021shape} which postprocesses the MMD of two consecutive sliding windows in order to find candidate drift points. The result of this postprocessing is the shape curve that indicates a candidate drift point when it changes sign from positive to negative. We illustrated the shape curve for different window lengths and one leakage in \cref{fig:shape}. As can be seen, a short window of just one day indicates drifts at the weekends, while for a window length of two weeks (or more), the drift induced by seasonal effects becomes stronger than the effect of the leak. Furthermore, for slightly less than a week the shape curve shows strong oscillations that correspond to the fact that we in fact have a weekly reoccurring pattern.


\section{Experiments\label{sec:experiments}}
For all our experiments, we use the data benchmark which we described in \cref{sec:data}. We will first investigate the suitability of model-loss-based drift detection and focus on data distribution-based drift detection later.\footnote{The experimental code is available at \url{https://github.com/FabianHinder/Drift-and-Water}}

\subsection{Model-Loss-Based Drift Detection}
\label{sec:model-loss}
\begin{figure*}[ht]
    \centering
    \includegraphics[width=\textwidth]{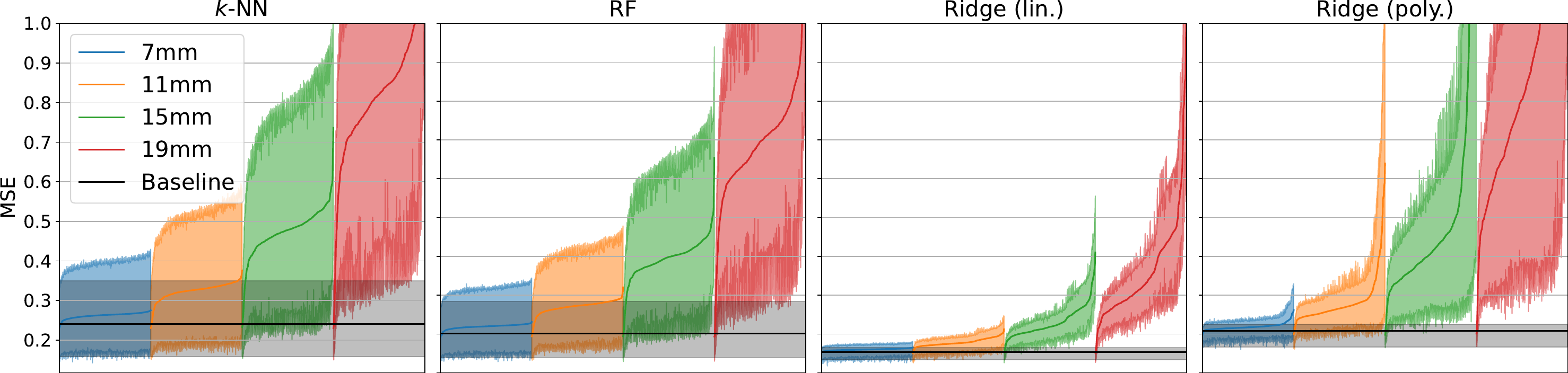}
    \caption{Mean squarred error of the forecasting task for different leakage sizes. Setting without leakage is reported as baselines. Line marks mean value, shaded area is minimal value to mean+standard deviation across folds.}
    \label{fig:forecast-mse}
\end{figure*}
\begin{figure*}[ht]
    \centering
    \includegraphics[width=\textwidth]{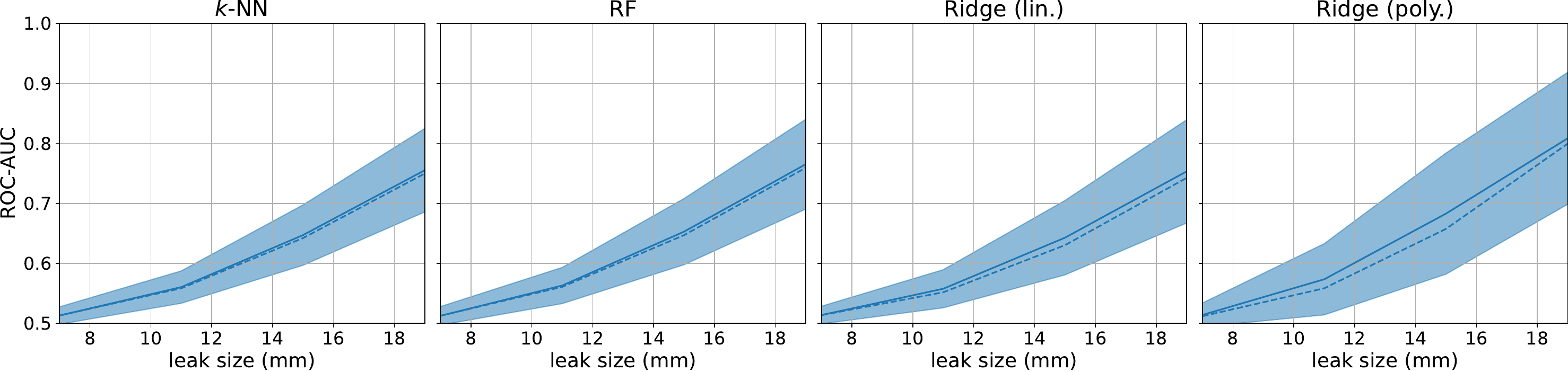}
    \caption{ROC-AUC for the model-loss-based drift detection for increasing leakagesizes. Forecasting task. Solid line marks mean value, shaded area mean$\pm$standard deviation, dashed line median.}
    \label{fig:forecast-auc}
\end{figure*}
To evaluate the model-loss-based detection schemes relying on forecasting and interpolation schemes for different regression models. Namely, we consider $k$NN, polynomial ridge regression, random forests, and linear ridge regression. RBF-ridge and RBF-/Poly-/Linear-SVR were considered but discarded after initial considerations due to weak performances in the regression task. In our experiments, we first analyze how well the models perform on the forecasting and interpolation task, and analyze their generalization capabilities to out-of-sample examples, e.g. to scenarios containing leakages. In the second step, we analyze how well the schemes are suited for detecting leakages. To do so we checked the underlying assumption that the model would perform better on the original training data (without leak) compared to the leaky data. In order for this to facilitate a useful strategy we need to be able to define a threshold $\theta$ such that $\text{MSE}(x_t) > \theta$ indicates a leakage at time $t$ and vice versa. Considering this as a classification problem with the classes ``no leak'' and ``leak'' we can apply the ROC AUC score to evaluate the performance of our models for leakage detection. In order to be more resilient to slowly growing leakages  we do not consider model updates. Thus, we end up with the following procedure:
\begin{enumerate}
    \item Select one fold: Extract two consecutive weeks from the baseline dataset
    \item Train the model on the data for the respective task
    \item Compute the errors of the model for the remaining year for each data point $E_0$
    \item Compute the errors of the model for the entire year for each leakagelocation and size $E_1$
    \item Compute the detection performance for this fold $\text{ROC-AUC}([0,\dots,0\;,\;1\dots,1], E_0 + E_1)$
\end{enumerate}
Recall that the ROC-AUC measures how well the obtained scores separate the leaky and non-leaky setups. The score is 1 if the largest error without leakage is smaller than the smallest error with leakage, it is 0.5 if the assignment is random. Thus, the ROC-AUC provides a scale-invariant upper bound on the performance of every concrete threshold. Furthermore, the ROC-AUC is not affected by class imbalance.

\paragraph{Forecasting}
\begin{figure}
    \centering
    \includegraphics[width=0.5\textwidth]{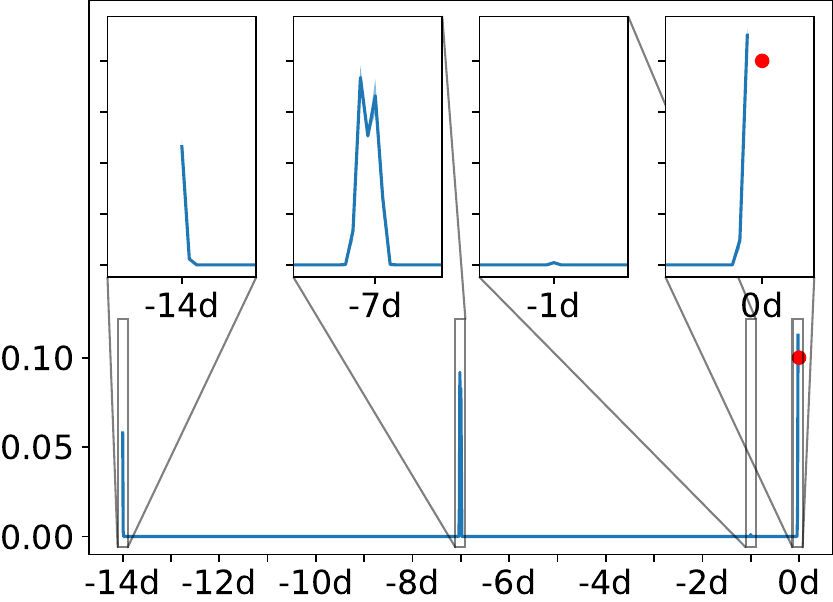}
    \caption{Mean absolute weight vector (and standard deviation) of forecasting task using ElasticNet. Features correspond to time points relative to the time point that is to be predicted (red dot).}
    \label{fig:relevance-profile}
\end{figure}

The results of our experiment evaluating the generalization ability of forecasting models are summarized in \cref{fig:forecast-mse}. We observe increasing errors for increasing leakage diameters. We find that our models generalize to small leakages, i.e. the MSE reported for the smallest leakage size is very similar to the baseline error on scenarios not containing any leakage. This might pose a problem for the detection of small leakages as the drift in the system is not reflected in the model loss. The small errors observed in the leakage-free scenarios indicate that the trained models can generalize well to the temporal patterns in the data.

\begin{figure*}[ht]
    \centering
    \includegraphics[width=\textwidth]{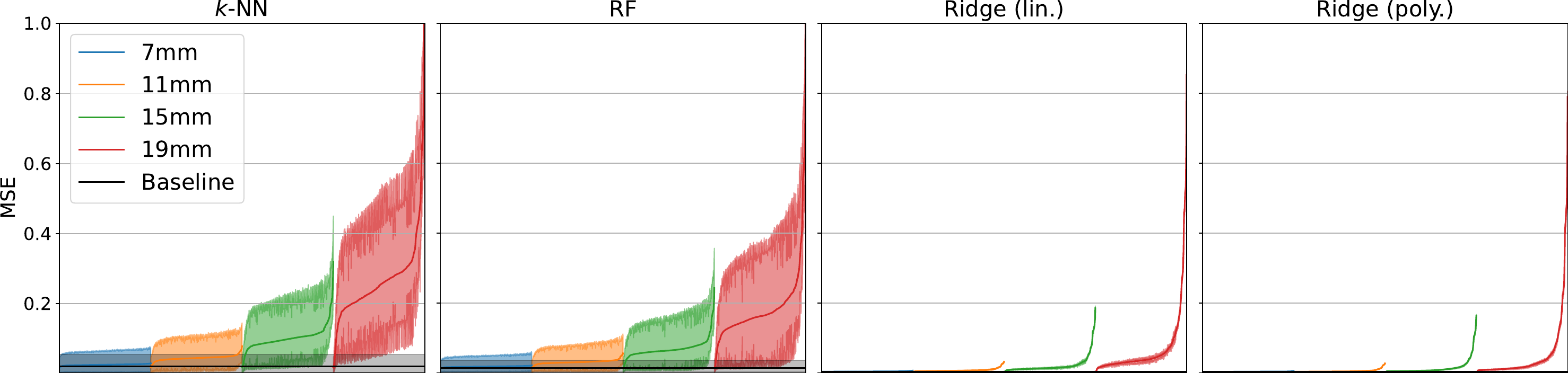}
    \caption{Mean squarred error of the interpolation task for different leakage sizes. Setting without leakage is reported as baselines. Line marks mean value, shaded area is minimal value to mean+standard deviation.}
    \label{fig:interpol-mse}
\end{figure*}

\begin{figure*}[ht]
    \centering
    \includegraphics[width=\textwidth]{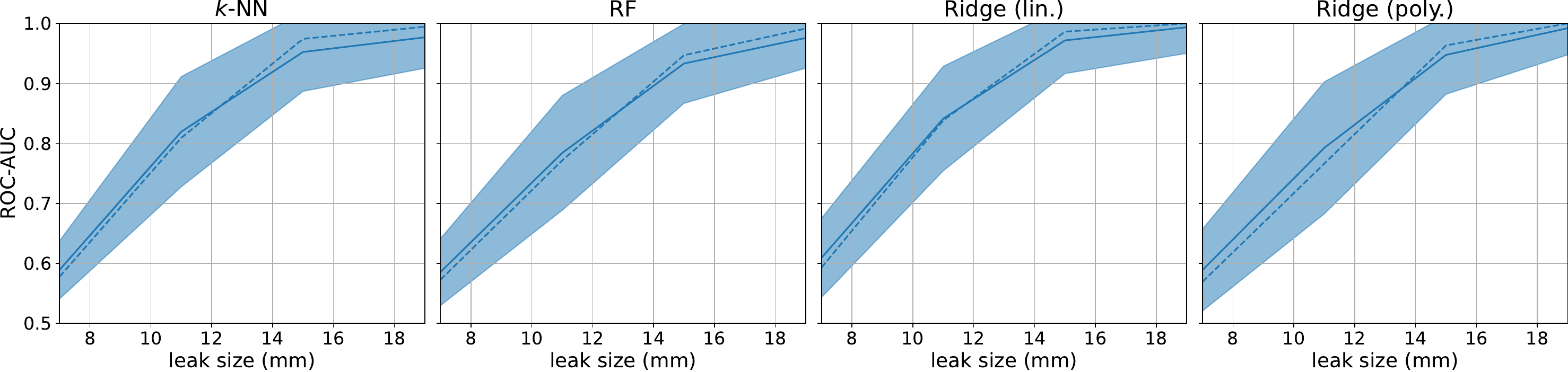}
    \caption{ROC-AUC for the model-loss-based drift detection for increasing leakage sizes. Interpolation task. Solid line marks mean value, shaded area mean$\pm$standard deviation, dashed line median.}
    \label{fig:interpol-auc}
\end{figure*}

Concerning the experiments on leakage detection, we report very low scores which align with random chance for small leakage sizes across all regression models. While the detection capabilities increase with increasing leakage size, the scores are still very weak.

As can be seen in \cref{fig:forecast-mse} linear models achieve comparably good performances. 
To get a better understanding of which features are important for the prediction we train an ElasicNet, a sparse linear model, and analyze its weight vector which is illustrated in \cref{fig:relevance-profile}. As can be seen, besides the value right before the value that is to be predicted, the values one and two weeks prior to it are very important. Furthermore, there is a slight relevance for one day prior to the event. All other values are zero, indicating that only those four time points are relevant for the prediction. This aligns well with our initial analysis of the temporal patterns in the data (\cref{sec:analysis}). We also repeated the same analysis with Ridge regression and Random Forest leading to similar results.

\paragraph{Interpolation}

The results of our experiment evaluating the generalization ability of the interpolation models are summarized in \cref{fig:interpol-mse}. We observe considerably lower errors than for the forecasting task. Again the errors increase with increasing leakage diameters and the models generalize to small leakages. In this setting, we find that the simple linear models perform much better than the $k$NN and the random forest. This aligns with findings published in \cite{vaquet_taking_2022} using simple linear models as virtual sensors.

The evaluation of the detection experiments is summarized in \cref{fig:interpol-auc}. In contrast to the forecasting setting, we observe reasonable ROC-AUC scores for leakage sizes of about 15mm-19mm. However, again, small leakages pose a difficult problem for model-loss-based drift detection schemes. Similar to the previous experiment, we observe scores that are only marginally above those that would be obtained by random guessing.

In summary, in our experiments, we find that model-loss-based detection schemes are only suitable for detecting large leakages. This aligns with both results of other works and the theory we briefly discussed in \cref{sec:drift}. In this use case the models generalize too well to out-of-distribution samples to reliably detect small leakages.

\subsection{Distribution-based Drift Detection}
\begin{figure*}[ht]
    \centering
    \includegraphics[width=\textwidth]{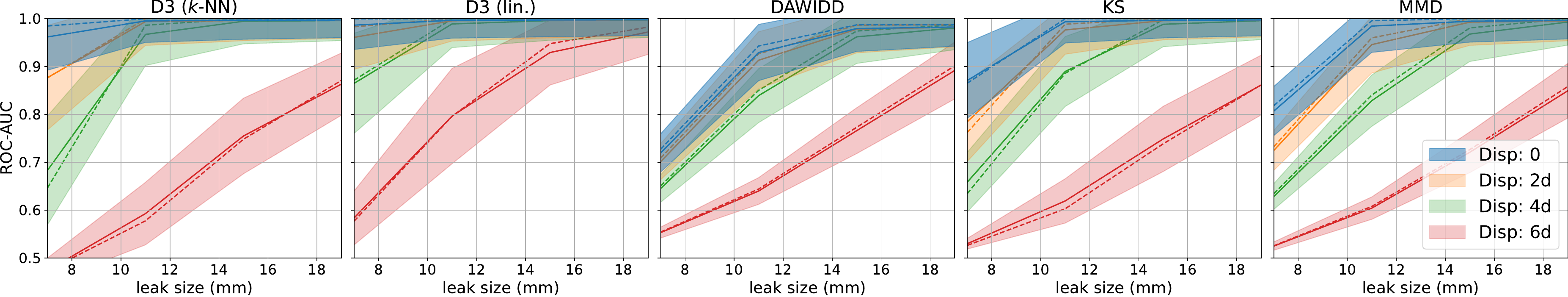}
    \caption{Performance of unsupervised drift detectors for different displacements (discrepancy between split point (assumed time-point of drift) and actual drift). Solid line marks mean value, shaded area mean$\pm$standard deviation, dashed line median.}
    \label{fig:dd-perf}
\end{figure*}
\begin{figure*}[ht]
    \centering
    \includegraphics[width=\textwidth]{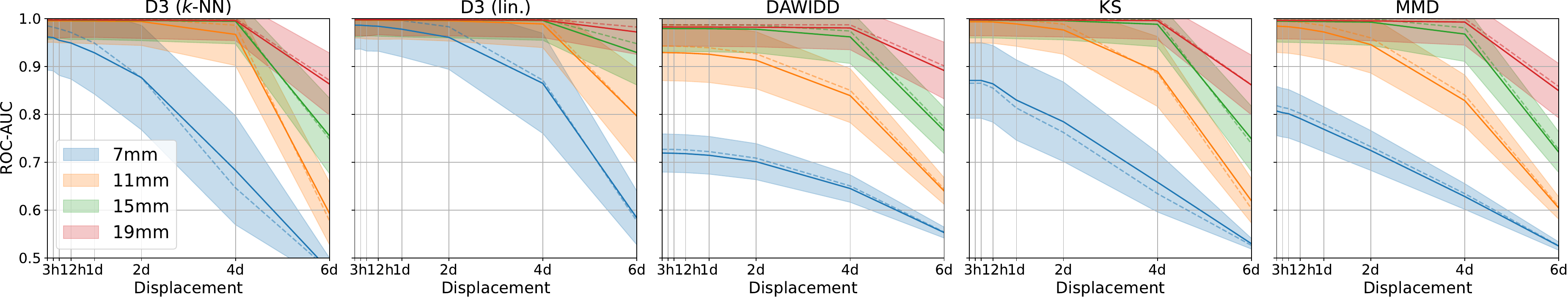}
    \caption{Performance of unsupervised drift detectors for different leakage sizes. Solid line marks mean value, shaded area mean$\pm$standard deviation, dashed line median.}
    \label{fig:dd-delay}
\end{figure*}

In this section, we describe our experiments with the different distribution-based drift detection schemes introduced in \cref{sec:distdd}. 
As argued before, choosing a suitable window size is crucial for the successful usage of distribution-based drift detection schemes. Analyzing the system at hand we decided to rely on two windows of one week each to eliminate the temporal patterns from the data. In addition to the choice of window size, the position of the split point, i.e. the border of the two considered windows, affects the performance of the detection scheme. The larger the displacement from the actual leakage onset time, the harder the detection will be. However, as it is desirable to detect leakages as soon as possible, it would be desirable to detect leakages even if the displacement is still large, e.g. if the split point still lies before the leakage occurred. If we get a reasonably high detection score, some kind of warning mechanism can be implemented where warnings can be confirmed when more data arrives.

In \cref{fig:dd-perf} we summarize the results of our distribution-based detection schemes with the different models. Assuming the split point lies exactly on the leakage, we report much better detection results than for the model-loss-based detection strategies. We report smaller scores for DAWIDD which is to be expected in this case as the model is not benefiting from our window size choice due to its block-based nature. Thus, it is affected by the temporal dependencies in the oscillations. While we obtain smaller scores for the statistical test-based methods for small leakages, the virtual classifier-based methods reliably detect leakages of a diameter of 7mm. 

As assumed, concerning the position of the split point, we observe a decline in the score with increasing split points. However, even for a displacement of 4 days, we obtain better scores than for the best model-based detection schemes. In the presence of a displacement of 6 days, we still get reasonable scores for large leakages when using the D3 detection scheme with a linear model. Thus detecting large leakages can be realized very fast, while for smaller leakages it takes a little more time to obtain a reliable warning.

These findings can be confirmed when analyzing \cref{fig:dd-delay}. One can see that apart from the smallest leakage size, with a displacement of 4 days, i.e. 3 days after the leakage occurred, most detection schemes yield a reasonable score, with the D3 variants performing best. Using the linear version of D3 even the smallest considered leakages can be detected with a delay of 5 days.

In conclusion, we found that distribution-based detection schemes outperform model-based detection if the window size is chosen in a suitable way. The methods are capable of detecting all leakage sizes. Large leakages can be detected with a reasonably small delay. In practical applications implementing a warning mechanism early on could be realized to react early on.

\subsection{Leakage localization}

Besides the task of leakage detection, i.e. determining whether or not there is a leakage somewhere in the WDN, there is also the more specific task of leakage localization, i.e. assuming there is a leakage somewhere determine the pipe where it occurrs~\cite{vrachimis_battle_2022}. This task is usually considered much harder and is commonly approached by trying to find the additional demand that is necessary to make the predicted pressures match the observed pressures\cite{li_fast_2022,daniel_sequential_2022,wang_multiple_2022,marzola_leakage_2022}. This of course again requires a lot of data usually not available to us. 
As our drift detection approach performed quite well on the detection task we consider the possibility to extend the methodology for leakage detection. Here the idea is quite simple the closer a sensor is to the leakage's actual position, the stronger the influence and thus the drift, leading to a particularly small $p$-value for that feature. As only the Kolmogorov-Smirnov test operates feature-wise we only take it into account. We make use of the same setup as before and narrow down the number of potential leakage locations by searching the sensor node closest to it. The closest sensor node is selected by choosing the one that has the smallest $p$-value, i.e. is considered to be particularly drifting by the test.

\newcommand{\dist}{Dist.}
\newcommand{\numCloser}{\#Cls.}
\newcommand{\relDist}{rel.D.}
In the following let $S$ be all sensor nodes, $s^*$ be the selected sensor node, and $v$ be the node where there leakage actually occurred, even if it is not the sensor node. Furthermore, $d$ denotes the graph distance in the WDN, i.e. $d(a,b)$ is the length of the shortest path connecting $a$ and $b$.
We make use of three metrics: distance between selected and actual node (\dist; $d(s^*,v)$), number of sensor nodes closer to the actual node (\numCloser; $|\{s \in S \mid d(s,v) < d(s^*,v)\}|$), and relative distance between actual node, selected and optimal node (\relDist; $d(s^*,v) / \min_{s \in S} d(s, v)$) which is normalized in contrast to the simple distance and smooth in contrast to the closer node metric.

\begin{table}[t]
    \centering
    \small
    \caption{\textsc{Results of leakage localization}}
    \begin{tabular}{c@{\quad}r@{$\pm$}l@{\quad}r@{$\pm$}l@{\quad}r@{$\pm$}l}
\toprule
size & \multicolumn{2}{l}{\dist} & \multicolumn{2}{l}{\relDist} & \multicolumn{2}{l}{\numCloser} \\
(mm) & $\mu$ & $\sigma$ & $\mu$ & $\sigma$ & $\mu$ & $\sigma$ \\
\midrule
7 & 10.1 & 13.1 & 2.6 & 4.9 & 3.3 & 7.0 \\
11 & 5.5 & 4.7 & 1.3 & 1.4 & 0.6 & 2.0 \\
15 & 5.1 & 3.9 & 1.2 & 1.2 & 0.5 & 1.5 \\
19 & 5.0 & 3.6 & 1.2 & 1.1 & 0.4 & 1.4 \\
\bottomrule
\end{tabular}
    \label{tab:loc}
\end{table}

The results are shown in \cref{tab:loc}. As can be seen, the results are quite promising. Furthermore, we observe that the precision is decreasing for smaller leakage sizes, which is to be expected considering the results from the last experiment. 

\section{Conclusion\label{sec:conclusion}}
In this work, we investigated the suitability of model-loss-based and distribution-based drift detection methods. Combining distribution-based detection with knowledge of WDNs, we provide detection schemes that successfully detect leakages of all sizes with reasonable detection delays. Analyzing model-loss-based techniques that are widely implemented in the water domain, we confirmed theoretical results that raise the issue of the loose connection between model loss and drift.

We assume that our work is not limited to WDNs but can also be realized for anomaly detection in other critical infrastructure systems like gas or electrical grids. In practical applications a further analysis of the leakages is necessary -- solely detecting leakages is not sufficient to take appropriate actions. We proposed a first localization strategy that is directly from detection efforts. Considering these follow-up tasks in more detail through the lens of concept drift is an interesting path for further research. Especially leakage localization requires further investigation as our initial results still leave space for improvement and require an extension to the actual leakage position not just the closest sensor node if applied in practice. A theoretical justification for this approach would be desirable.

\section*{{Acknowledgements}}

We gratefully acknowledge funding from the European Research Council (ERC) under the ERC Synergy Grant Water-Futures (Grant agreement No. 951424).

\bibliographystyle{apalike}
{\small
\bibliography{example}}

\begin{thebibliography}{}

\bibitem[{ European Commission}, 2021]{EU_proposal_AI}
{ European Commission} (2021).
\newblock {Artificial Intelligence Act}.

\bibitem[Cardell-Oliver and Carter-Turner, 2021]{smart-meters}
Cardell-Oliver, R. and Carter-Turner, H. (2021).
\newblock Activity-aware privacy protection for smart water meters.
\newblock In {\em 8th ACM BuildSys}, BuildSys '21, page 31–40. Association for Computing Machinery.

\bibitem[Daniel et~al., 2022]{daniel_sequential_2022}
Daniel, I., Pesantez, J., Letzgus, S., Khaksar~Fasaee, M.~A., Alghamdi, F., Berglund, E., Mahinthakumar, G., and Cominola, A. (2022).
\newblock A {Sequential} {Pressure}-{Based} {Algorithm} for {Data}-{Driven} {Leakage} {Identification} and {Model}-{Based} {Localization} in {Water} {Distribution} {Networks}.
\newblock {\em J. Water Resour. Plan. Manag.}, 148.

\bibitem[Eggimann et~al., 2017]{eggimann_potential_2017}
Eggimann, S., Mutzner, L., Wani, O., Schneider, M.~Y., Spuhler, D., Moy~de Vitry, M., Beutler, P., and Maurer, M. (2017).
\newblock The {Potential} of {Knowing} {More}: {A} {Review} of {Data}-{Driven} {Urban} {Water} {Management}.
\newblock {\em Environmental Science \& Technology}, 51(5):2538--2553.

\bibitem[Eliades and Polycarpou, 2010]{eliades_fault_2010}
Eliades, D.~G. and Polycarpou, M.~M. (2010).
\newblock A {Fault} {Diagnosis} and {Security} {Framework} for {Water} {Systems}.
\newblock {\em IEEE Transactions on Control Systems Technology}, 18(6):1254--1265.

\bibitem[G{\"o}z{\"u}a{\c{c}}{\i}k et~al., 2019]{d3}
G{\"o}z{\"u}a{\c{c}}{\i}k, {\"O}., B{\"u}y{\"u}k{\c{c}}ak{\i}r, A., Bonab, H., and Can, F. (2019).
\newblock Unsupervised concept drift detection with a discriminative classifier.
\newblock In {\em Proceedings of the 28th ACM international conference on information and knowledge management}, pages 2365--2368.

\bibitem[Gretton et~al., 2006]{mmd}
Gretton, A., Borgwardt, K.~M., Rasch, M.~J., Sch{\"{o}}lkopf, B., and Smola, A.~J. (2006).
\newblock A kernel method for the two-sample-problem.
\newblock In {\em NIPS 2006}, pages 513--520.

\bibitem[Gretton et~al., 2007]{hsic}
Gretton, A., Fukumizu, K., Teo, C.~H., Song, L., Sch{\"{o}}lkopf, B., and Smola, A.~J. (2007).
\newblock A kernel statistical test of independence.
\newblock In {\em NIPS 2007}, pages 585--592.

\bibitem[Hinder et~al., 2020]{hinder2020dynamic}
Hinder, F., Artelt, A., and Hammer, B. (2020).
\newblock Towards non-parametric drift detection via dynamic adapting window independence drift detection {(DAWIDD)}.
\newblock In {\em ICML 2020}, volume 119, pages 4249--4259. {PMLR}.

\bibitem[Hinder et~al., 2021]{hinder2021shape}
Hinder, F., Brinkrolf, J., Vaquet, V., and Hammer, B. (2021).
\newblock A shape-based method for concept drift detection and signal denoising.
\newblock In {\em 2021 IEEE Symposium Series on Computational Intelligence (SSCI)}, pages 01--08. IEEE.

\bibitem[Hinder et~al., 2023a]{hinder_change_2023}
Hinder, F., Vaquet, V., Brinkrolf, J., and Hammer, B. (2023a).
\newblock On the {Change} of {Decision} {Boundary} and {Loss} in {Learning} with {Concept} {Drift}.
\newblock In {\em IDA {XXI}}, volume 13876, pages 182--194. Springer Nature Switzerland, Cham.

\bibitem[Hinder et~al., 2023b]{hinder_hardness_2023}
Hinder, F., Vaquet, V., Brinkrolf, J., and Hammer, B. (2023b).
\newblock On the {Hardness} and {Necessity} of {Supervised} {Concept} {Drift} {Detection}:.
\newblock In {\em 12th ICPRAM}, pages 164--175, Lisbon, Portugal. SCITEPRESS.

\bibitem[Hinder et~al., 2023c]{hinder2023things}
Hinder, F., Vaquet, V., and Hammer, B. (2023c).
\newblock One or two things we know about concept drift -- a survey on monitoring evolving environments.

\bibitem[Hu et~al., 2018]{hu_survey_2018}
Hu, C., Li, M., Zeng, D., and Guo, S. (2018).
\newblock A survey on sensor placement for contamination detection in water distribution systems.
\newblock {\em Wireless Networks}, 24(2):647--661.

\bibitem[Kolomogorov, 1933]{an1933sulla}
Kolomogorov, A. (1933).
\newblock Sulla determinazione empirica di una legge didistribuzione.
\newblock {\em Giorn Dell'inst Ital Degli Att}, 4:89--91.

\bibitem[Lambert, 1994]{lambert_accounting_1994}
Lambert, A. (1994).
\newblock Accounting for {Losses}: {The} {Bursts} and {Background} {Concept}.
\newblock {\em Water and Environment Journal}, 8(2):205--214.

\bibitem[Laucelli et~al., 2016]{laucelli_detecting_2016}
Laucelli, D., Romano, M., Savić, D., and Giustolisi, O. (2016).
\newblock Detecting anomalies in water distribution networks using {EPR} modelling paradigm.
\newblock {\em Journal of Hydroinformatics}, 18(3):409--427.

\bibitem[Li et~al., 2022]{li_fast_2022}
Li, Z., Wang, J., Yan, H., Li, S., Tao, T., and Xin, K. (2022).
\newblock Fast {Detection} and {Localization} of {Multiple} {Leaks} in {Water} {Distribution} {Network} {Jointly} {Driven} by {Simulation} and {Machine} {Learning}.
\newblock {\em J. Water Resour. Plan. Manag.}, 148(9).

\bibitem[Mala-Jetmarova et~al., 2018]{mala-jetmarova_lost_2018}
Mala-Jetmarova, H., Sultanova, N., and Savic, D. (2018).
\newblock Lost in {Optimisation} of {Water} {Distribution} {Systems}? {A} {Literature} {Review} of {System} {Design}.
\newblock {\em Water}, 10(3):307.

\bibitem[Marzola et~al., 2022]{marzola_leakage_2022}
Marzola, I., Mazzoni, F., Alvisi, S., and Franchini, M. (2022).
\newblock Leakage {Detection} and {Localization} in a {Water} {Distribution} {Network} through {Comparison} of {Observed} and {Simulated} {Pressure} {Data}.
\newblock {\em J. Water Resour. Plan. Manag.}, 148(1):04021096.

\bibitem[Rodell et~al., 2018]{rodell2018emerging}
Rodell, M., Famiglietti, J.~S., Wiese, D.~N., Reager, J., Beaudoing, H.~K., Landerer, F.~W., and Lo, M.-H. (2018).
\newblock Emerging trends in global freshwater availability.
\newblock {\em Nature}, 557(7707):651--659.

\bibitem[Romano et~al., 2014]{romano_automated_2014}
Romano, M., Kapelan, Z., and Savić, D.~A. (2014).
\newblock Automated {Detection} of {Pipe} {Bursts} and {Other} {Events} in {Water} {Distribution} {Systems}.
\newblock {\em J. Water Resour. Plan. Manag.}, 140(4):457--467.

\bibitem[Romero-Ben et~al., 2022]{romero-ben_leak_2022}
Romero-Ben, L., Alves, D., Blesa, J., Cembrano, G., Puig, V., and Duviella, E. (2022).
\newblock Leak {Localization} in {Water} {Distribution} {Networks} {Using} {Data}-{Driven} and {Model}-{Based} {Approaches}.
\newblock {\em J. Water Resour. Plan. Manag.}, 148(5).

\bibitem[Rossman, 2000]{rossman_epanet_2000}
Rossman, L.~A. (2000).
\newblock {EPANET} 2: users manual.
\newblock US Environmental Protection Agency. Office of Research and Development.

\bibitem[Steffelbauer et~al., 2022]{steffelbauer_pressure-leak_2022}
Steffelbauer, D.~B., Deuerlein, J., Gilbert, D., Abraham, E., and Piller, O. (2022).
\newblock Pressure-{Leak} {Duality} for {Leak} {Detection} and {Localization} in {Water} {Distribution} {Systems}.
\newblock {\em J. Water Resour. Plan. Manag.}, 148(3).

\bibitem[Vaquet et~al., 2022]{vaquet_taking_2022}
Vaquet, V., Artelt, A., Brinkrolf, J., and Hammer, B. (2022).
\newblock Taking {Care} of {Our} {Drinking} {Water}: {Dealing} with {Sensor} {Faults} in {Water} {Distribution} {Networks}.
\newblock In {\em {ICANN} 2022}, volume 13530, pages 682--693. Springer Nature Switzerland, Cham.

\bibitem[V{\"o}r{\"o}smarty et~al., 2010]{vorosmarty2010global}
V{\"o}r{\"o}smarty, C.~J., McIntyre, P.~B., Gessner, M.~O., Dudgeon, D., Prusevich, A., Green, P., Glidden, S., Bunn, S.~E., Sullivan, C.~A., Liermann, C.~R., et~al. (2010).
\newblock Global threats to human water security and river biodiversity.
\newblock {\em nature}, 467(7315):555--561.

\bibitem[Vrachimis et~al., 2022]{vrachimis_battle_2022}
Vrachimis, S.~G., Eliades, D.~G., Taormina, R., Kapelan, Z., Ostfeld, A., Liu, S., Kyriakou, M., Pavlou, P., Qiu, M., and Polycarpou, M.~M. (2022).
\newblock Battle of the {Leakage} {Detection} and {Isolation} {Methods}.
\newblock {\em Journal of Water Resources Planning and Management}, 148(12).

\bibitem[Wang et~al., 2022]{wang_multiple_2022}
Wang, X., Li, J., Liu, S., Yu, X., and Ma, Z. (2022).
\newblock Multiple {Leakage} {Detection} and {Isolation} in {District} {Metering} {Areas} {Using} a {Multistage} {Approach}.
\newblock {\em J. Water Resour. Plan. Manag.}, 148(6).

\end{thebibliography}



\end{document}